\documentclass[12 pt]{article}
\usepackage{authblk}
\usepackage{graphicx}
\usepackage{float}
\usepackage[verbose=true,letterpaper]{geometry}
\AtBeginDocument{
	\newgeometry{
		textheight=9in,
		textwidth=5.5in,
		top=1in,
		headheight=12pt,
		headsep=25pt,
		footskip=30pt
	}

}

\usepackage[sorting=none, backend=biber]{biblatex}
\title{Improved Forward-Forward Contrastive Learning}
\author{Gananath R\\
\texttt{gananathr@gmail.com}
}

\date{}

\begin{document}

\maketitle

\begin{abstract}
\noindent The backpropagation algorithm, or backprop, is a widely utilized optimization technique in deep learning. While there's growing evidence suggesting that models trained with backprop can accurately explain neuronal data, no backprop-like method has yet been discovered in the biological brain for learning. Moreover, employing a naive implementation of backprop in the brain has several drawbacks. In 2022, Geoffrey Hinton proposed a biologically plausible learning method known as the Forward-Forward (FF) algorithm. Shortly after this paper, a modified version called FFCL was introduced. However, FFCL had limitations, notably being a three-stage learning system where the final stage still relied on regular backpropagation. In our approach, we address these drawbacks by eliminating the last two stages of FFCL and completely removing regular backpropagation. Instead, we rely solely on local updates, offering a more biologically plausible alternative.
\end{abstract}

\section{Introduction}

Artificial Neural Networks (ANNs) draw inspiration from the structure and function of the human brain. These networks comprise interconnected nodes called neurons, organized into layers. ANNs possess the remarkable ability to learn from complex data, adapt their structure, and make predictions or decisions based on their learning process. The groundwork for ANNs began in the 1940s and 1950s when researchers like Warren McCulloch and Walter Pitts proposed simplified mathematical models for neurons \cite{McCulloch1943}. In 1957, Rosenblatt's perceptron laid the foundation for neural network theory. Rosenblatt categorized his perceptron models into two types based on the input they received: photoperceptrons for processing images or visual data, and phonoperceptrons for handling sound inputs. By introducing simplified models inspired by the human brain, Rosenblatt demonstrated the potential of artificial neurons to perform computational tasks \cite{Rosenblatt1957,Rosenblatt1958}.

The rise and widespread adoption of ANNs today can largely be attributed to the introduction of the backpropagation algorithm, often referred to simply as backprop. This algorithm is a fundamental principle within the field of ANNs, enabling machines to learn and adapt quickly and effectively based on data. The term "backpropagation" was first mentioned by Rosenblatt in 1962 when he proposed a method for propagating error corrections back to the sensory end of the network \cite{Rosenblatt1962}. However, it was the work of Rumelhart in 1986 that significantly contributed to the popularization of the backpropagation algorithm and accelerated its acceptance and adaptation within the research community.

Backpropagation embodies the principle of iterative refinement, allowing neural networks to progressively enhance their predictive capabilities. This algorithm comprises two stages: forward propagation, also known as inference or prediction, and backward propagation, which facilitates learning. During forward propagation, inputs from the preceding layer are passed to the subsequent layer while undergoing transformations. As the input data traverses the network, it undergoes successive transformations until reaching the output layer, where a prediction is produced. In the backward phase, an error value is computed using a loss function, typically comparing the predicted value with the ground truth. Gradients are then calculated with respect to the weight parameters using the error. The objective of backpropagation is to update these weight parameters of the neural network using the gradients in a manner that minimizes the error \cite{Rumelhart1986}.

Backprop stands as a cornerstone in the advancement of deep learning and artificial intelligence. Alongside strides in hardware, algorithms, and data accessibility, it has been pivotal in driving applications like OpenAI's ChatGPT, showcasing remarkable abilities in generating human-like text. While there is mounting evidence supporting the explanatory power of backprop-trained models in interpreting neural response data, direct evidence of a backprop-like algorithm being employed by the brain for learning remains elusive. Several obstacles hinder its direct implementation in biological neural systems, such as the absence of local error representation, the requirement for synaptic symmetry in both forward and backward pathways, and the necessity for signed error signals, among others. These challenges pose significant hurdles and underscore the complexity of bridging artificial and biological learning mechanisms \cite{Lillicrap2020,Whittington2019}.

In the scientific community, there's a growing interest in finding a biologically plausible alternative to backpropagation. In 2022, Hinton proposed the Forward-Forward (FF) algorithm as one such alternative, drawing inspiration from Boltzmann machines and Noise Contrastive Estimation. Unlike backpropagation, which employs both forward and backward passes, FF relies on two separate forward passes. One pass involves processing positive data, while the other processes negative data. Each layer in the FF model is equipped with its own loss functions. The goal of FF algorithm is to boost the activity within layers for positive data while reducing it for negative data \cite{Hinton2022}.

Within a few months of Hinton's paper, Ahmed et al. published a short paper titled "Forward-Forward Contrastive Learning". In their paper, Ahmed et al. introduced a novel contrastive training method termed the FFCL algorithm. The FFCL algorithm comprises a modified FF algorithm with three distinct learning stages, with the initial two serving as pretraining phases. During the first stage, known as local contrastive learning, they execute local updates for individual blocks. Following this, they proceed to a global update phase aimed at learning global representations. The final stage employs regular backpropagation for conducting downstream classification tasks. Despite its efficacy, a notable drawback of FFCL is its continued reliance on backpropagation, along with the computational intensity associated with its three-stage training process \cite{Ahamed2023}.

Recently Aghagolzadeh and colleagues have introduced a method akin to FFCL, dubbed Forward-Forward with Marginal Supervised Contrastive loss (FFCM). Unlike FFCL's three-stage training process, FFCM condenses it into two stages. In the first stage, FFCM employs a comparable strategy to train encoder layers as FFCL does. Subsequently, in the second stage, FFCM integrates a classification layer onto the frozen encoder and proceeds with additional training using cross entropy loss for a limited number of epochs \cite{Aghagolzadeh2024}. While FFCM represents an improvement over FFCL, it continues to rely on a two-stage training process. In contrast, our method operates within a single stage, rendering it significantly more efficient than FFCM.

\section{Proposed Method}

\begin{figure}[t]
	\includegraphics[scale=0.3]{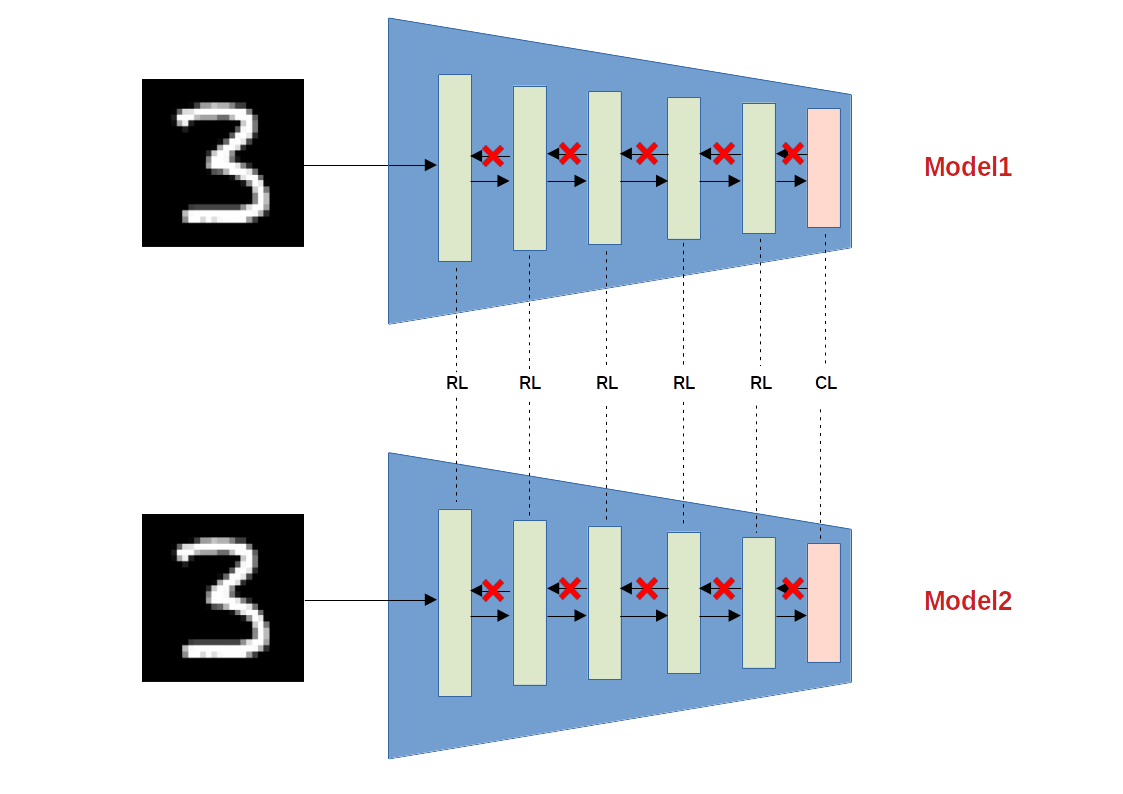} 
	\caption{This figure serves as a visual representation of our proposed model.}
	\label{fig:fullmodel}
\end{figure}

In our proposed technique, we have omitted the second and third stages of the original FFCL algorithm. Instead, we utilize two separate instances of the same model, each with the same number and type of layers, including the output classification layer. The weights in the trainable layers are randomly initialized for the two different models. Each set of corresponding trainable layers also has a common loss function, which is utilized for error computation, gradient calculation, and weight update within that layer. Since our method lacks a global scaler representation to monitor the overall performance of the network, we employed the output from one model's layers as a guiding tool to train the corresponding layer in the second model. 

We will follow the same training approach that is used in the local contrastive representation learning stage (first stage) in FFCL's method except that output from last hidden layer will be passed to the classification layer and this layer is also involved in training. For local updates we have used Representation Loss (RL) for all trainable layers except last output layer were Classification Loss (CL) has been used. Figure \ref{fig:fullmodel} shows the structure of our trainable model. There is no backpropagation through the layers in any part of our model, that is, weights are updated locally only.

\section{Experiments and Results}
\begin{figure}[t]
	\includegraphics[scale=0.2]{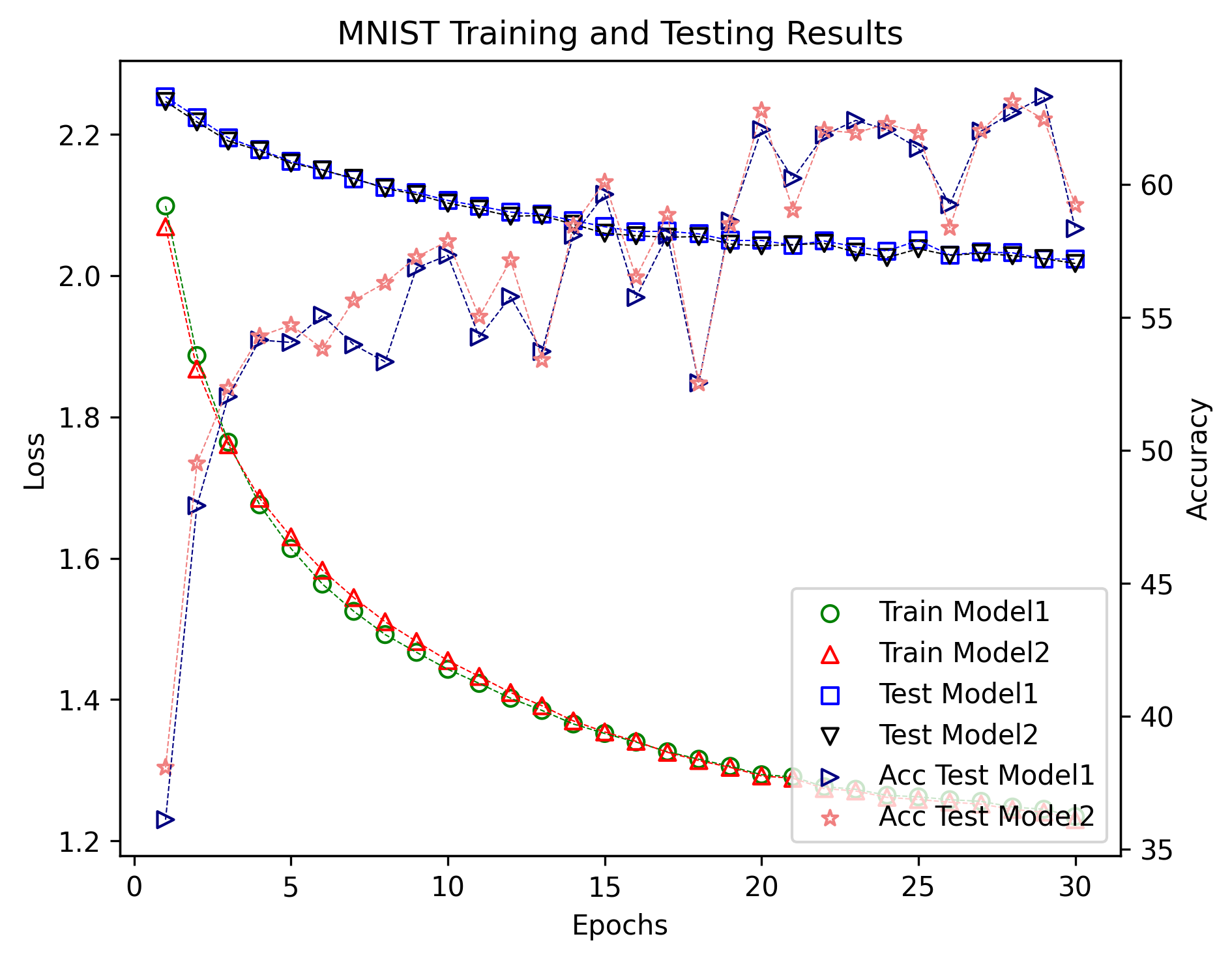} 
	\caption{The plot illustrates the train and test classification loss values, along with the test accuracies, for the two models.}
	\label{fig:result}
\end{figure}

The model comprises of four linear layers: an input layer with 784 nodes, two hidden layers with 64 nodes each, and a 10-node output layer. All layers utilize ReLU activations except for the final layer, which employs softmax activation. Cosine embedding loss serves as the representation loss for non-classification layers, while binary cross-entropy loss is applied as the classification loss for the last layer. We utilized the Adam optimizer with a learning rate of 0.0001 for weight optimization and trained it for 30 epochs. The MNIST training dataset was leveraged for training purposes, while the test set is used for gathering metrics at the end of each epoch. Positive sampling of data was exclusively carried out, and input images from the dataset are passed through two distinct models concurrently. As the signals propagate forward, the corresponding layers of both models learn simultaneously, obviating the need for backpropagation through layers.

Figure \ref{fig:result} shows the training and testing loss values for the last layer, along with the testing accuracies obtained from our study. Both training and testing phases exhibited an exponential decay in losses, indicative of successful ANN training. Initially, the testing accuracy for both models started near 37\% and eventually reached as high as 63\% during training. The experiments were conducted multiple times, and Figure \ref{fig:result} represents the results of one such experiment. All other experiments conducted in this study yielded similar outcomes (data not included).

\subsection{Biological Plausibility}
The Hebbian theory posits that neurons that fire together wire together. In both human and animal brains, there exists a class of neurons known as mirror neurons. These neurons activate both when an individual performs an action or when they observe someone else doing the same action. Studies on mirror neurons have revealed that they encode not only muscle movements or joint changes but also the goals of actions \cite{Umilta2008}. Furthermore, research indicates that the mirror neuron system aids motor learning by enhancing the execution of training movements during imitation-based motor learning \cite{Stefan2005}. In our proposed method, we utilize the output from one model's layer as a guiding tool for learning in the corresponding layer of the second model. Although we don't assert this as definitive proof of the biological feasibility of our proposed model, we highly encourage further research to substantiate this hypothesis. 

\section{Conclusion}
In this paper, we address enhancements to the FFCL algorithm. One of the primary drawbacks of the FFCL algorithm lies in its utilization of a three-stage training process, wherein the third stage relies on backpropagation. This approach not only increases the algorithm's complexity but also escalates computational demands during training. Our proposed solution streamlines the FFCL algorithm into a single stage and eliminates the need for backpropagation entirely.

\subsection*{Disclaimer}
This study was conducted independently by the author, utilizing personal time and resources. No external funding or resources were utilized in its completion.

\printbibliography
\end{document}